%% file: main.tex
\documentclass{article} 
\usepackage{iclr2026_conference,times}

\input{math_commands.tex}

\usepackage{hyperref}
\usepackage{url}
\usepackage{graphicx}
\usepackage{booktabs}
\usepackage{multirow}
\iclrfinalcopy

\usepackage{fontawesome}
\usepackage{xspace}

\title{\textsc{\textsc{\textsc{MedGPT-oss}}}: Training a General-Purpose Vision-Language Model for Biomedicine}

\author{Kai Zhang$^{1,\dagger}$, Zhengqing Yuan$^{2}$, Cheng Peng$^{3}$, Songlin Zhao$^{1}$, Mengxian Lyu$^{3}$, Ziyi Chen$^{3}$, \\
\textbf{Yanfang Ye$^{2}$, Wei Liu$^{4}$, Ying Zhang$^{5}$, Kaleb E Smith$^{6}$, Lifang He$^{1}$, Lichao Sun$^{1,*}$} \textbf{, Yonghui Wu$^{3,}$}\thanks{Corresponding authors.} \\
\\
$^{1}$Department of Computer Science and Engineering, Lehigh University \\
$^{2}$Department of Computer Science and Engineering, University of Notre Dame \\
$^{3}$Department of Health Outcomes \& Biomedical Informatics, University of Florida \\
$^{4}$Department of Radiation Oncology, Mayo Clinic\\
$^{5}$Research Computing, University of Florida\\
$^{6}$AI Technology Center, NVIDIA\\
$^{\dagger}$Current affiliation: Department of Radiation Oncology, Stanford University.\\
\texttt{lis221@lehigh.edu, yonghui.wu@ufl.edu}
}

\begin{document}

\maketitle

\begin{abstract}
Biomedical multimodal assistants have the potential to unify radiology, pathology, and clinical-text reasoning, yet a critical deployment gap remains: top-performing systems are either closed-source or computationally prohibitive, precluding the on-premises deployment required for patient privacy and PHI compliance. We introduce \textsc{MedGPT-oss}, an open-weight, 20B-parameter generalist vision-language model designed to facilitate open research in clinical AI. Rather than relying on architectural complexity, \textsc{MedGPT-oss} pairs the GPT-oss language backbone with a visual front-end via a optimized, three-stage training curriculum. By progressively domain-adapting these modules through rigorous data curation and long-context multimodal alignment, we demonstrate that a 20B model can bridge the capacity gap. It successfully outperforms larger open medical models on out-of-distribution (OOD) multimodal reasoning and complex text-only clinical tasks. By unifying diverse modalities under a single instruction-following interface, \textsc{MedGPT-oss} maintains a parameter-efficient footprint fully compatible with commodity GPUs. We release the complete training recipe, open-weight checkpoints, and a rigorous evaluation harness to serve as a verifiable foundation for privacy-preserving, institution-specific clinical AI research.
\end{abstract}


\section{Introduction}

Biomedical AI has made rapid progress across image interpretation, clinical-text understanding, and multimodal reasoning \cite{thirunavukarasu2023large}. Yet, most deployed systems remain specialists: models trained for a single task and often a single modality \cite{moor2023foundation}. This specialization complicates real-world clinical use, where decision-making is inherently multimodal and longitudinal, requiring the continuous integration of radiographs, pathology slides, and patient narratives over time. 

Recent ``generalist'' vision-language models (VLMs) aim to unify these heterogeneous inputs within a single backbone. However, the most capable systems are predominantly closed-source, precluding the on-premises deployment required to satisfy strict safety auditing and data privacy (PHI) mandates. While open-weight medical VLMs have emerged to address this \cite{llava_med, zhang2024generalist}, the field faces a delicate trade-off between scaling for complex reasoning and maintaining practical deployability. Smaller models (e.g., 7B) often lack the capacity for rigorous multi-step clinical logic, whereas larger models present elevated serving costs and memory bottlenecks for standard IT infrastructure. Furthermore, many current approaches assume that bridging this gap requires initializing from computationally expensive, pre-existing domain-specific vision encoders.

In this technical report, we introduce \textsc{MedGPT-oss}, an open-weight, 20B-parameter generalist biomedical VLM designed to address the challenge of balancing advanced clinical reasoning with accessible deployment. Starting with the robust GPT-oss language backbone \cite{agarwal2025gpt} and a standard CLIP \cite{radford2021learning} visual encoder connected via a simple linear projector, we establish that architectural minimalism---when supported by an optimized, progressive training curriculum---is remarkably effective. By continuously updating the visual and projection weights alongside the language model during our alignment and mid-training phases, we demonstrate that a systematic data recipe can dynamically adapt general-purpose modules to the medical domain, achieving frontier-level multimodal alignment without relying on bespoke architectural engineering.

\textsc{MedGPT-oss}-20B demonstrates highly competitive performance across multiple challenging benchmarks. Notably, it matches or exceeds the performance of larger open-sourced models in nearly half of our evaluations (8 of 17) focusing on complex, multi-step clinical reasoning and multimodal in-context learning. Furthermore, it exhibits a robust ability to resist negative transfer when provided with unseen clinical demonstrations. This parameter-efficient footprint ensures the model is fully compatible with commodity GPUs, enabling seamless on-premises fine-tuning and inference. In summary, our contributions are as follows:

\begin{itemize}
    \item \textbf{A unified, multi-stage training recipe:} We detail a fully reproducible, three-stage curriculum (short-context alignment, long-context mid-training, and mixed instruction tuning) that progressively domain-adapts general-purpose vision and language modules. This pipeline proves that strategic data curation and joint parameter updating can empower a 20B model to punch significantly above its weight class, eliminating the strict dependency on specialized vision encoders.
    \item \textbf{SOTA empirical performance:} \textsc{MedGPT-oss}-20B establishes new state-of-the-art results among open models on challenging OOD benchmarks (e.g., MedFrameQA, MedXQA). Notably, it crosses the reasoning threshold required for complex clinical QA, outperforming 30B+ systems in both zero-shot and few-shot generalization.
    \item \textbf{Accessible clinical deployment:} By releasing our open-weight checkpoints, comprehensive data recipes, and standardized evaluation harnesses, we provide a complete ecosystem for on-premises, air-gapped clinical deployment. This enables healthcare institutions to conduct rigorous, privacy-preserving validation without exporting sensitive patient data.
\end{itemize}

This report positions \textsc{MedGPT-oss} not as a finalized clinical oracle, but as a highly capable, transparent foundation for the medical AI research community to build upon, audit, and deploy safely. Our ongoing work focuses on developing larger models and integrating advanced techniques to further the development of open, reliable medical AI.

\section{Related Work}

\paragraph{Specialist pipelines and their limits.}
Early biomedical AI largely advanced through modality- and task-specific pipelines. In radiology, report generation commonly followed encoder-decoder designs that map images to long-form text, but generation quality is often sensitive to supervision biases and prone to omission or hallucination when summarizing complex findings \citep{johnson2019mimic}. Medical VQA developed with task-specific models that frequently reduce answering to classification over constrained spaces, limiting expressivity and robustness under distribution shift \citep{Lau2018VQARAD,Liu2021SLAKE}. In computational pathology, weakly supervised multiple-instance learning became a standard recipe for whole-slide understanding, but it is typically vision-only and tightly coupled to slide-level supervision, making cross-task transfer and multimodal grounding difficult \citep{Lu2021CLAM}. Clinical NLP similarly evolved as text-only models tailored to individual supervised objectives (e.g., clinical NLI), which often generalize poorly beyond their training distributions \citep{Romanov2018MedNLI}. Across these silos, recurring bottlenecks include limited cross-task transfer, weak multimodal integration, and unreliable long-form generation \citep{yang2022large}.

\paragraph{Generalist biomedical VLMs and instruction-tuned assistants.}
A key inflection point was the move from per-task systems toward generalist biomedical vision-language modeling. Representation- and retrieval-oriented pretraining (e.g., large-scale biomedical image-text contrastive learning) improved transfer across diverse downstream tasks and enabled stronger cross-modal alignment \citep{zhang2025multimodal}. Generalist architectures that unify heterogeneous biomedical tasks under a single model further reduced task fragmentation and improved cross-task reuse \citep{zhang2024generalist, peng2025scaling}. In parallel, domain instruction tuning of general-purpose multimodal assistants demonstrated that a strong base model can be specialized into a medically useful interactive assistant with relatively lightweight post-training \citep{llava_med}. Beyond biomedical-first models, unified multimodal backbones (e.g., sequence-to-sequence formulations) provided effective starting points for medical adaptation. Together, these lines shifted emphasis from designing task-specific heads to building reusable multimodal interfaces and training recipes that support broader coverage and transfer.

\paragraph{Frontier trends: recipes, long-context grounding, and evaluation.}
Recent work pushes generalist medical MLLMs further by treating \emph{training and evaluation recipes} as first-class: multi-stage curricula that combine continued pretraining with instruction tuning, higher-quality and more diverse medical mixtures (including synthetic instructions/reasoning traces), longer-context multimodal grounding, and stronger reliability-oriented evaluation toolkits and protocols \citep{jiang2025hulu,xu2025lingshu,sellergren2025medgemma,ossowski2025octomed,dai2025qoqmed}. Closed systems often report strong zero-/few-shot performance and long-context capabilities, but limited disclosure constrains reproducibility and independent auditing \citep{Tu2023,OpenAI2023}. Overall, the field is converging on a performance--transparency--deployability frontier, where progress increasingly depends on reproducible recipes and standardized, reliability-aware evaluation rather than benchmark scores alone.

\paragraph{Positioning of \textsc{\textsc{MedGPT-oss}}.}
\textsc{\textsc{MedGPT-oss}} targets a practical open-weight regime motivated by privacy and governance constraints in clinical deployment: open weights and permissive licensing support on-premises and even air-gapped use, enabling institution-specific adaptation without exporting PHI. 
To ensure reproducibility and reliable comparison, we disclose training recipes and evaluation code, prevent train--test overlap to avoid leakage, and standardize benchmarking via a unified inference harness with deterministic decoding and strict, fully automated scoring rules.
Methodologically, we adopt a simple three-stage curriculum (short alignment $\rightarrow$ long-context mid-training $\rightarrow$ mixed multimodal/text instruction tuning).
Finally, we emphasize reliability-oriented evaluation by combining standardized automatic scoring with task-specific, clinically informed assessment protocols, aligning model development with externally valid reporting.

\section{\textsc{\textsc{MedGPT-oss}}}
In the following, we delineate the design of \textsc{\textsc{\textsc{MedGPT-oss}}} across three fundamental pillars: (a) model architecture, (b) training strategies, and (c) implementation details (e.g., hyperparameter selection).

\subsection{Model Architecture}\label{sec:architecture}
\textsc{\textsc{\textsc{MedGPT-oss}}} follows a modular multimodal architecture aligned with recent vision–language models such as LLaVA~\citep{liu2023visual, an2025llava} and Qwen-VL~\citep{bai2023qwen, Qwen3-VL} series. Our model is composed of three core components: (1) a visual encoder for extracting image features; (2) a projection module that maps visual embeddings into the language space; and (3) a language backbone adapted from open-weight large language models.

\paragraph{\textbf{Visual Encoder.}} Following established paradigms in medical MLLMs \citep{llava_med,chen-etal-2024-towards-injecting,xu2025lingshu,jiang2025hulu}, we employ the \texttt{CLIP-ViT-L/14@336px} \citep{radford2021learning} as our primary vision backbone. This architecture was selected to strike an optimal balance between granular representational capacity and computational efficiency during inference. To bridge the modality gap, input medical images ranging from radiographs to histopathology slides are mapped into a high-dimensional latent space and subsequently aligned with the LLM’s embedding space via a projection layer.

\begin{figure}
    \centering
    \includegraphics[width=0.8\linewidth]{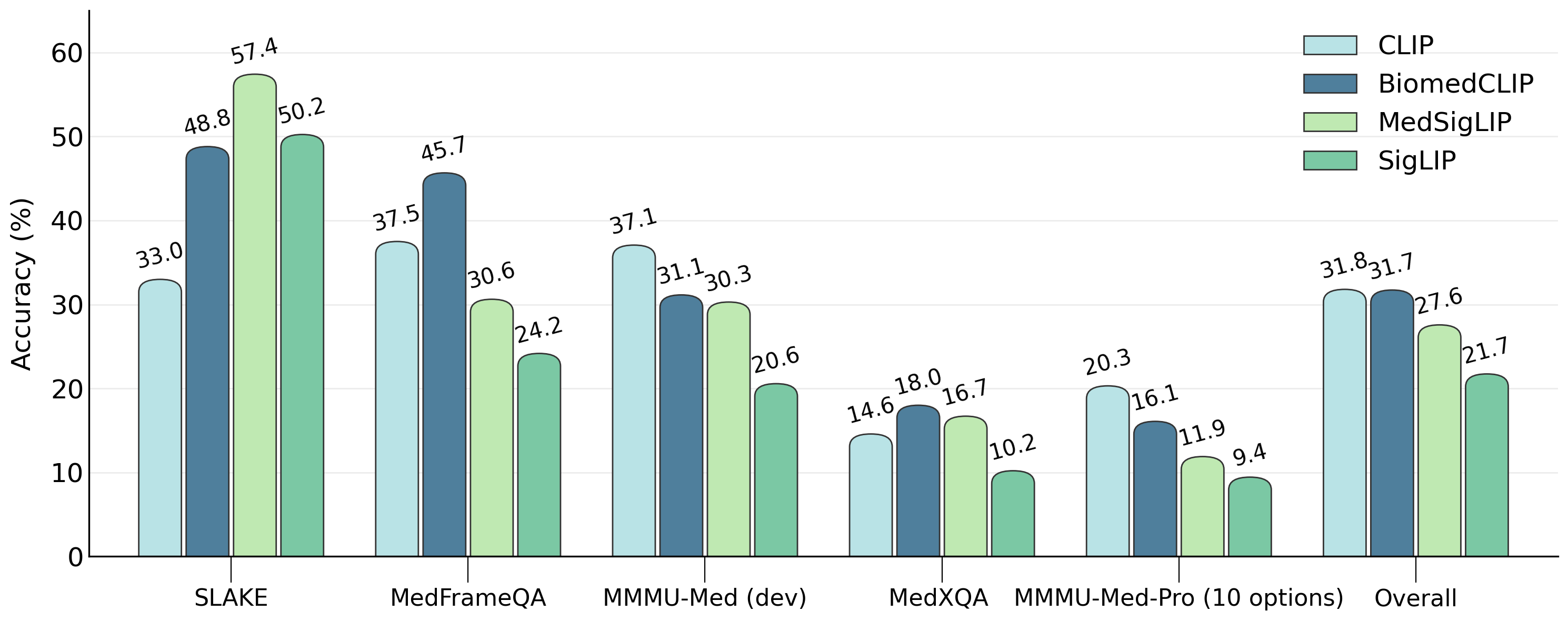}
    \caption{Preliminary evaluation of visual encoders on medical multimodal benchmarks. As an initial investigation, we compared the vanilla CLIP backbone against domain-specific alternatives (BiomedCLIP, MedSigLIP) and SigLIP. The models utilize a GPT-oss-20B trained via LoRA (2-stage training following vanilla LLaVA, where we first pretrain the projector, then fine-tune the projector and LLM backbone).}
    \label{fig:vision_encoder_ablation}
\end{figure}

Although domain-specific encoders are tailored for clinical contexts, our selection of the vanilla CLIP backbone is substantiated by empirical benchmarking as shown in Figure \ref{fig:vision_encoder_ablation}. Specifically, we conducted a comparative ablation study utilizing a 20B-parameter model variant trained via low-rank adaptation (LoRA) \citep{hu2022lora}. This evaluation leveraged the comprehensive BiomedGPT training corpus \citep{zhang2024generalist}, comprising 1 million multimodal pairs and 9 billion tokens of biomedical text. Our results demonstrate that vanilla CLIP consistently outperformed specialized alternatives, including BiomedCLIP \citep{zhang2025multimodal}, SigLIP \citep{zhai2023sigmoid}, and MedSigLIP \citep{sellergren2025medgemma}, suggesting that the broader visual priors captured during large-scale general pretraining may provide a more resilient foundation for downstream medical reasoning. 

\paragraph{\textbf{Projection Module.}}
Architectural choices for the projection module typically follow two primary paradigms: (1) Bottlenecked attention mechanisms, such as the Q-Former \citep{li2023blip, carion2020end}, which minimize trainable parameters but may inadvertently impair spatial grounding due to the information loss inherent in fixed-length query pooling and unstable training dynamics \citep{yao2024deco}; and (2) Direct projection layers \citep{liu2023visual, yuan2023tinygpt}. We opt for the latter, employing a two-layer multi-layer perceptron (MLP) that directly maps CLIP visual features into the language model's input embedding space.

\paragraph{\textbf{LLM Backbone.}} For the linguistic core of \textsc{\textsc{\textsc{MedGPT-oss}}}, we employ GPT-oss in its 20B (full parameters) 
configuration \cite{agarwal2025gpt}. These models were selected for their state-of-the-art performance on clinical reasoning benchmarks, notably surpassing comparable open-weight models in factual grounding and domain-specific coherence \cite{arora2025healthbench}. 

Beyond their specialized capacity for high-fidelity medical instruction-following, these variants represent an optimal trade-off within the current scaling landscape. The 20B configuration facilitates high-performance deployment on commodity hardware.
By utilizing these backbones, \textsc{\textsc{\textsc{MedGPT-oss}}} retains robust general-purpose capabilities, ensuring it remains a versatile MLLM across diverse clinical scenarios. Furthermore, this open-weight architecture democratizes access to elite medical AI, empowering clinical institutions to maintain local data control and reducing reliance on high-resource cloud infrastructure.

\subsection{Training Strategies}\label{sec:training_strategies}

We implement a multi-stage or curriculum learning pipeline designed to progressively bridge the domain gap between general-purpose vision-language representations and specialized clinical requirements. All data of training phase are preventing overlap with evaluation data and avoiding potential leakage, details in Table \ref{tab:training_data}.

\begin{table}[ht]
    \centering
    \caption{Overview of \textsc{\textsc{\textsc{MedGPT-oss}}} training datasets.}
    \label{tab:training_data}
    \renewcommand{\arraystretch}{1.2}
    \resizebox{\textwidth}{!}{
    \begin{tabular}{l|l|r|c|p{7.5cm}}
        \toprule
        \textbf{Modality} & \textbf{Dataset} & \textbf{No. examples} & \textbf{Training stages} & \textbf{Description} \\
        \midrule
        \multirow{15}{*}{Text-only} 
        & \href{https://huggingface.co/datasets/Intelligent-Internet/II-Medical-Reasoning-SFT}{II-Medical-Reasoning} & 2,197,741 & IT & (R) Clinical dialogues \& QA pairs\\
        & \href{https://huggingface.co/datasets/BAAI/Infinity-Instruct}{Infinity-Instruct-Gen} & 1,483,097 & IT & General high-quality chat instructions \\
        & \href{https://huggingface.co/datasets/lingshu-medical-mllm/ReasonMed}{ReasonMed} & 1,111,555 & IT & (R) Medical QA data \\
        &
        \href{https://huggingface.co/datasets/openlifescienceai/medmcqa}{MedMCQA} & 182,822 & IT & Indian medical entrance exam questions\\
        & \href{https://huggingface.co/datasets/lavita/ChatDoctor-HealthCareMagic-100k}{HealthCareMagic-100k} & 112,165 & IT & Real conversations between patients and doctors \\
        & \href{https://huggingface.co/datasets/GBaker/MedQA-USMLE-4-options}{MedQA} & 10,178 & IT & USMLE style exam questions \\
        & \href{https://huggingface.co/datasets/HiTZ/MedExpQA}{MedExpQA} & 434 & IT & Spanish medical residency exam questions \\
        & \href{https://huggingface.co/datasets/qiaojin/PubMedQA}{PubMedQA} & 1,000 & IT & Biomedical QA compiled from PubMed abstracts \\
        & \href{https://huggingface.co/datasets/lavita/AlpaCare-MedInstruct-52k}{AlpaCare-MedInstruct-52k} & 52,002 & IT & Synthetic medical instruction data curated from 167 clinical-craft tasks \\
        & \href{https://huggingface.co/datasets/lavita/MedQuAD}{MedQuAD} & 47,441 & IT & Medical QA pairs created from 12 NIH websites \\
        & \href{https://huggingface.co/datasets/FreedomIntelligence/medical-o1-verifiable-problem}{Huotuo-o1} & 40,644 & IT &  (R) Verifiable QA pairs from exam questions \\
        & \href{https://huggingface.co/datasets/UCSC-VLAA/MedReason}{MedReason} & 32,682 & IT & (R) Clinical QA with structured reasoning\\
        & \href{https://huggingface.co/datasets/FreedomIntelligence/Medical-R1-Distill-Data}{Medical-R1-Distill-Data} & 22,000 & IT & (R) Medical reasoning distilled from Deepseek-R1\\
        & \href{https://huggingface.co/datasets/hw-hwei/MedThoughts-8K}{MedThoughts-8K} & 7,716 & IT & (R) Distilled reasoning based on MedQA data \\
        & \href{https://huggingface.co/datasets/zhengComing/iCliniq-10K}{iCliniq-10K} & 7,321 & IT & Real patient-doctor conversation \\
        \hline
        \multirow{6}{*}{Radiology} 
        & \href{https://github.com/sctg-development/ROCOv2-radiology}{ROCOv2 (filtered)} & 38,861 & PT & Radiological images and associated medical captions \\
        & \href{https://physionet.org/content/mimic-cxr-jpg/2.1.0/}{MIMIC-CXR (single-panel)} & 22,747 & MT & Chest X-ray images \& free-form reports \\
        & \href{https://github.com/taokz/EditGRPO}{MIMIC-CXR (multi-view \& longitudinal)} & 146,893 & MT & Chest X-ray images \& free-form reports \\
        & \href{https://github.com/baeseongsu/mimic-cxr-vqa}{MIMIC-CXR-VQA} & 363,598 & IT & Chest X-ray images \& VQA pairs \\
        & \href{https://huggingface.co/datasets/BoKelvin/SLAKE}{SLAKE (English)} & 9,835 & IT & 2D CT/MRI/chest X-ray images \& VQA pairs \\
        & \href{https://huggingface.co/datasets/flaviagiammarino/vqa-rad}{VQA-Rad} & 1,793 & IT & Radiology image QA pairs \\
        &\href{https://huggingface.co/datasets/BoKelvin/GEMeX-ThinkVG}{GEMeX-ThinkVG} & 202,383 & IT & (R) Chest X-ray images \& grounded VQA pairs \\
        \hline
        \multirow{2}{*}{Histopathology}
        & \href{https://github.com/wisdomikezogwo/quilt1m}{Quilt-1M (filtered)} & 359,725 & PT & Histopathology image-caption pairs\\
        & \href{https://huggingface.co/datasets/wisdomik/Quilt_VQA}{Quilt-VQA} & 107,131 & IT & Histopathology VQA pairs curated from videos \\
        \hline
        \multirow{4}{*}{Mixture} 
        &\href{https://huggingface.co/datasets/axiong/pmc_oa}{PMC-OA (filtered)} & 477,078 & PT & Image-caption pairs from PMC Open Access Subset \\
        &\href{https://huggingface.co/BIOMEDICA/datasets}{BIOMEDICA (dermatology and surgery subsets)} & 63,775 & PT & Image-caption pairs from PMC Open Access Subset \\
        &\href{https://huggingface.co/datasets/UCSC-VLAA/MedTrinity-25M}{MedTrinity (accessible)} & 18,525,974 & MT & Medical images \& enriched annotations/captions \\
        & \href{https://huggingface.co/datasets/FreedomIntelligence/PubMedVision}{PubMedVision (alignment)} & 646,759 & MT &  Image-caption pairs extracted from PubMed articles \\
        &\href{https://huggingface.co/datasets/RadGenome/PMC-VQA}{PMC-VQA} & 188,229 & IT & Curated VQA pairs from PMC Open Access Subset \\
        & \href{https://huggingface.co/datasets/baichuan-inc/OpenMM_Medical}{OpenMM-Medical} & 88,996 & IT & Multiple-domain multi-choice VQA pairs \\
        & \href{https://huggingface.co/datasets/General-Medical-AI/GMAI-Reasoning10K}{GMAI-Reasoning10K} & 9,965 & IT & (R) Medical image reasoning covering 12 imaging modalities \\
        \bottomrule
        \multicolumn{5}{l}{\textbf{PT:} Pretraining, \textbf{MT:} Mid-training, \textbf{IT:} Instruction-tuning. \textbf{(R):} Reasoning data} \\
    \end{tabular}
    }
\end{table}

\paragraph{Pretraining.}
We first pretrain on short medical image--text pairs to establish robust cross-modal alignment between heterogeneous medical imaging modalities and their corresponding textual descriptions, providing a foundation for subsequent medical knowledge integration. During this initial adaptation phase, we freeze the parameters of the LLM backbone to preserve its foundational linguistic and reasoning capabilities. Conversely, we keep the visual encoder and the projection module trainable. 

The pretraining corpus is constructed from PMC-OA \citep{lin2023pmc}, Quilt-1M \citep{ikezogwo2023quilt}, ROCOv2 \citep{ruckert2024rocov2}, and BIOMEDICA (dermatology and surgery subsets) \citep{lozano2025biomedica}. To ensure high-quality, clinically relevant supervision, we restrict to single-panel images and filter out common artifacts, including visible narrators, desktop environments, pathology software interfaces, and overlaid text within images. We further remove clinically irrelevant pairs and misleading numeric content with the assistance of large language models. In total, the resulting pretraining set contains approximately 939K pairs.

\paragraph{Mid-training.} We then perform mid-training on comprehensive, long-context image--text pairs to more fully integrate medical knowledge into the MLLM, improving its ability to recognize diverse medical concepts and generalize across clinical contexts. During this stage, we relax all architectural constraints and enable full-parameter updates across the visual encoder, projection module, and LLM backbone.

For this stage, we use MIMIC-CXR \citep{johnson2019mimic} training set (real-world radiology reports' findings section), the open-sourced subset of MedTrinity \citep{xie2025medtrinitym}, and the PubMedVision \citep{chen-etal-2024-towards-injecting} Alignment dataset. Since MedTrinity and PubMedVision are already curated and verified, we apply no additional preprocessing to these sources. For MIMIC-CXR, we retain single-image report pairs and additionally curate longitudinal study sets following previous works \citep{zhang2025editgrpo, zhang2025libra}. The mid-training corpus comprises approximately 19M samples, which is inspired by LLaVA-OneVision-1.5 \citep{an2025llava} positing that scaling high-quality data at the mid-training stage can yield state-of-the-art multimodal performance.

\paragraph{Instruction-tuning.} In line with prevailing practices for MLLMs, we include a medical instruction-tuning stage to strengthen the model’s instruction-following as well as reasoning ability across a broad range of task directives, thereby aligning its outputs more closely with user intent. Similar to mid-training, full parameters are activated.

To ensure diversity, we curate a mixed corpus that combines medical visual instruction-tuning data with pure language instruction-tuning datasets (with and without reasoning trajectories). The resulting instruction-tuning set contains approximately 7M pairs. Note that the reasoning trajectories are generally distilled from state-of-the-art LLMs through reject sampling and refinement operations. We select multiple versions based on different datasets and models to provide diverse reasoning trajectories.

\subsection{Implementation Details}
\label{sec:implementation}

The training hyperparameters are shown in Table~\ref{tab:hparams}. Specifically, we trained the \textsc{MedGPT-oss}-20B medical instruction model using a three-stage recipe on 8 $\times$ NVIDIA B200 GPUs, employing DeepSpeed \citep{rasley2020deepspeed} ZeRO-3 to enable efficient distributed optimization. Unless otherwise specified, we optimized the model using AdamW \citep{loshchilov2018decoupled} ($\beta_1=0.9$, $\beta_2=0.999$, $\epsilon=10^{-8}$) with a weight decay of 0.05. We applied a cosine learning rate decay schedule with a warmup ratio between 0.01 and 0.03, utilizing bfloat16 precision and gradient checkpointing throughout. 

Training was conducted with a maximum sequence length of 32{,}768. We used a per-device batch size of 16, resulting in a global batch size of 128 across the 8 GPUs (with a gradient accumulation step of 1). For long-context adaptation, we integrated YaRN RoPE \citep{peng2024yarn} scaling (scaling factor of 32, $\theta=150{,}000$) to extend the original context length of 4{,}096 to 131{,}072 positions.

For visual inputs, we enabled dynamic image sizing supporting up to 12 dynamic patches, enforced a 336-pixel image size with a downsampling ratio of 0.5, and included thumbnails in the visual preprocessing pipeline. Stage-wise wall-clock training times on the B200 cluster were 8 hours and 27 minutes for pretraining (1 epoch), 240 hours and 24 minutes for mid-training (1 epoch), and 86 hours and 26 minutes for instruction-tuning (1 epoch).

\begin{table}[ht]
\centering
\small
\caption{Default implementation hyperparameters. LR = learning rate.}
\label{tab:hparams}
\resizebox{0.7\textwidth}{!}{
\begin{tabular}{ll}
\toprule
Hyperparameter & Default value \\
\midrule
Optimizer & AdamW (DeepSpeed ZeRO-3) \\

LR schedule & Cosine decay with warmup ratio 0.01–0.03 \\

Max sequence length & 32{,}768 \\

Per-device batch size & 16 (global batch size: 128) \\

Projector LR (stage 0) & $1\times10^{-3}$ (LLM frozen) \\

LLM LR (mid-training) & $2\times10^{-5}$ \\

LLM LR (instruction tuning) & $8\times10^{-5}$ \\

Vision encoder & Frozen in stage 0–1; jointly trained in stage 2 \\

Weight decay & 0.05 \\

Precision & bfloat16 with gradient checkpointing \\
\bottomrule
\end{tabular}
}
\end{table}

\section{Experiments}\label{sec:experiments}
\subsection{Datasets}
We comprehensively evaluate \textsc{\textsc{\textsc{MedGPT-oss}}} across core capability dimensions commonly reported in recent medical foundation model studies, covering multimodal reasoning, biomedical knowledge, report generation, and in-context learning. The details of each data are listed in Table \ref{tab:eval_data}.

\paragraph{\textbf{Visual Question Answering (VQA).}} 
We benchmark multimodal medical reasoning on six representative datasets using their official evaluation splits unless otherwise specified: SLAKE \citep{Liu2021SLAKE}, MedFrameQA \citep{yu2025medframeqa}, MMMU-Med (dev) \citep{yue2024mmmu}, MMMU-Pro-Med (4- and 10-option test sets) \citep{yue2025mmmu}, and MedXQA (multimodal) \citep{zuo2025medxpertqa}. The MMMU(-Pro)-Med benchmark is constructed by aggregating the clinically relevant subject subsets from MMMU(-Pro), including \textit{basic medical science, clinical medicine, diagnostics and laboratory medicine, pharmacy, and public health}.

To ensure fair and reproducible evaluation, we standardize all tasks into closed-ended multiple-choice formats where applicable. This avoids reliance on LLM-as-a-Judge protocols, which are known to introduce bias and evaluation instability \citep{ye2025justice}. All results are therefore computed via exact-match accuracy under structured answer constraints.

\begin{table}[h]
\centering
\caption{Overview of \textsc{\textsc{MedGPT-oss}} evaluation datasets.}
\label{tab:eval_data}
\resizebox{0.9\textwidth}{!}{%
\begin{tabular}{llrrc}
\toprule
\textbf{Task} & \textbf{Dataset} & \textbf{No. Images} & \textbf{No. Examples} & \textbf{OOD$^\dagger$} \\ \midrule
\multirow{6}{*}{VQA} & MedXQA (multimodal) & 1.43 ± 1.04  & 	2,000  & \checkmark \\
 & SLAKE (English) & 1.00 ± 0.00  & 836  & - \\
 & MedFrameQA & 3.24 ± 1.26 & 2,851  & \checkmark \\ 
 & MMMU-Med (dev) & 1.05 ± 0.29 & 175 & \checkmark \\
 & MMMU-Med-Pro (4 or 10 options) & 1.14 ± 0.57 & 286 & \checkmark \\
 \hline

\multirow{6}{*}{Text QA} & MedQA & - & 1,273 & - \\
 & MedMCQA & - & 4,183 & - \\
 & PubMed QA & - & 500 & - \\
 & MMLU-Med & - & 1,395 & - \\
 & MedXQA (text-only) & - & 2,450 & \checkmark \\
 & Medbullets & - & 	298 & - \\
 \hline

Report generation & MIMIC-CXR (multi-view \& longitudinal) & 4.30 ± 0.62 & 2,231 & - \\ \hline
\multirow{2}{*}{In-context learning $^*$} 
 & Patient-trial & - & 4,936 & \checkmark \\ 
 & Impression generation & 3.39 ± 0.64 & 1,564 & \checkmark \\
 \bottomrule
\multicolumn{5}{p{\linewidth}}{\footnotesize $^\dagger$ OOD: Out of Distribution; $^*$ Data statistics are derived from the few-shot configurations.} \\
\end{tabular}%
}
\end{table}

\paragraph{\textbf{Text-only Question Answering.}} 
We assess clinical and biomedical reasoning using MedQA \citep{jin2021disease}, PubMedQA \citep{jin2019pubmedqa}, MedMCQA \citep{pal2022medmcqa}, MedXQA (text-only), MMLU-Med \citep{hendrycks2021measuring}, and MedBullets \citep{chen2025benchmarking}. MMLU-Med integrates seven MMLU subsets: \textit{anatomy, clinical knowledge, college biology, college medicine, medical genetics, nutrition, and professional medicine}. Consistent with the VQA evaluation protocol, all tasks are curated into standardized multiple-choice settings to enable objective scoring and eliminate generative evaluation ambiguity.

\paragraph{\textbf{Report Generation.}} 
For structured medical generation, we evaluate chest X-ray findings generation on the MIMIC-CXR benchmark (including multi-view and longitudinal settings) \citep{zhang2025editgrpo}. We report three preliminary metrics to assess clinical correctness and linguistic quality.

\paragraph{\textbf{In-Context Learning (ICL).}} 
Beyond standard zero-shot benchmarking, we explicitly evaluate the in-context learning capability of \textsc{\textsc{\textsc{MedGPT-oss}}}, measuring its ability to leverage limited task-specific demonstrations to generalize to unseen or unfamiliar medical tasks. We report both zero-shot and few-shot performance.

For multimodal ICL,  
we curate a chest X-ray impression generation task based on MIMIC-CXR that is not included in the model’s training corpus, enabling evaluation of out-of-distribution task adaptation. We construct the dataset by pairing each study’s radiograph(s) with its corresponding reference \textit{Impression} section, joining image records and textual annotations via study identifiers. Each instance is formulated as an image-conditioned generation task, optionally augmented with structured contextual metadata to guide reporting. To ensure modality–label consistency, we exclude studies with missing impressions or unavailable image files. For in-context evaluation, we further create a 1-shot variant by prepending a randomly sampled, label-valid exemplar to the query input. The demonstration–query structure is fixed under a controlled random seed to ensure reproducibility.

We further incorporate patient–trial matching benchmarks derived from the SIGIR 2016 Clinical Trials corpus \citep{koopman2016test} and the 2021 and 2022 TREC Clinical Trials Tracks \citep{soboroff2021overview, roberts2022overview}, evaluating the model’s ability to reason over structured clinical context for eligibility determination under minimal supervision. For each patient query, we pair the patient note with retrieved candidate trials and retain only patient–trial pairs with official relevance judgments in the released qrels. Relevance grades are mapped to eligibility labels following a deterministic scheme (2/1/0 $\rightarrow$ eligible/partially eligible/not eligible). To improve interpretability and evidence localization, patient notes are standardized via sentence segmentation with explicit sentence identifiers. Trial documents are transformed into structured prompts containing the trial title, target diseases, interventions, summary, and cleaned inclusion/exclusion criteria (with headers, bullet/number prefixes, and trivial fragments removed). Each retained patient–trial pair is formatted as a multiple-choice eligibility question. The answer set consists of four fixed options: \textit{eligible}, \textit{partially eligible}, \textit{not eligible}, and \textit{not enough information}. Option ordering is randomized per instance under a fixed deterministic seed to ensure reproducibility.

\subsection{Baselines}

We compare \textsc{\textsc{\textsc{MedGPT-oss}}-20B} against most recent open-source MLLMs, with an emphasis on size-matched baselines to ensure fair comparison at the ~20B+ scale. These baselines span diverse backbone families and design paradigms in medical AI.

\textbf{MedGemma-27B} \citep{sellergren2025medgemma} is trained for medical text and image comprehension using a SigLIP \citep{zhai2023sigmoid} vision encoder pre-trained on de-identified medical datasets, and demonstrates strong baseline performance on clinically relevant benchmarks including question answering and report generation.

\textbf{Lingshu-32B} \citep{xu2025lingshu} is an open-source medical MLLM built on the Qwen2.5-VL backbone, specialized via a staged ``shallow-to-deep'' medical adaptation pipeline: vision--language medical alignment on curated image--text pairs, followed by medical instruction tuning (and an optional reinforcement learning variant) to strengthen clinically grounded multimodal reasoning across diverse imaging modalities.

\textbf{Hulu-Med-32B} is a generalist medical vision--language model that combines a SigLIP-family vision encoder with a Qwen-based language decoder, bridged by a multimodal projector and trained via a progressive multimodal curriculum to support medical understanding across 2D images, 3D volumes, videos, and clinical text \citep{jiang2025hulu}.

We also include additional open medical models such as \textbf{OctoMed-7B} \citep{ossowski2025octomed} and \textbf{QoQ-Med-32B} \citep{dai2025qoqmed}, which represent alternative design trade-offs in training data composition and multimodal fusion strategies. 

Collectively, these baselines cover a spectrum of architectural choices (e.g., Gemma-based, Qwen-based, and other backbones) and multimodal training paradigms, providing a comprehensive context for evaluating the merits and limitations of \textsc{\textsc{\textsc{MedGPT-oss}}}.

\subsection{Inference and Scoring}

\textbf{All results reported in this paper are generated using our own inference harness} with publicly released checkpoints; no numbers are copied from original publications due to the data difference. To ensure standardized and reproducible evaluation, we (i) convert each benchmark into a unified chat-style prompt format, (ii) apply the official chat template for each model when available, and (iii) implement dataset-specific post-processing and scoring rules consistent with their original definitions. 

\paragraph{\textbf{Decoding.}}
For baseline models, we replicate the official inference guidelines and publicly released codebases, including recommended hyperparameters and decoding configurations. Unless otherwise specified, we perform a single deterministic decode per example on all medical benchmarks, matching the evaluation philosophy adopted by the original works. No sampling-based reruns or majority voting are used.

\paragraph{\textbf{Multiple-choice scoring.}}
For multiple-choice tasks, we employ rule-based option extraction under structured answer constraints. Prompts explicitly specify answer options, and predictions are considered correct only if they exactly match one of the valid choices. Outputs that do not correspond to any predefined option are counted as incorrect, as instruction adherence is treated as part of the model’s capability. This strict exact-match protocol avoids subjective interpretation and ensures fair, fully automated scoring.

\paragraph{\textbf{Chest X-ray report assessment.}} Evaluating the quality of generated radiology reports is inherently challenging: widely used general-purpose NLP metrics (e.g., n-gram overlap or generic semantic similarity) often penalize harmless paraphrases while failing to adequately capture clinically critical errors such as missed findings or incorrect negation. As a result, the field has increasingly adopted clinically informed evaluation metrics that aim to better reflect medical correctness and utility. In this work, we use the following metrics for report quality measurement: 
\textbf{RadGraph-F1} \citep{jain2021radgraph} \footnote{By default, we utilize the ``partial'' reward level to allow for partial credit based on semantic overlap, offering a more nuanced evaluation than strict exact matching.} considers literal entity agreement considering the positive or negative context of each entity. \textbf{RadCliQ} \citep{yu2023evaluating} \footnote{The original RadCliQ metric is designed to be lower-is-better. To ensure consistency with other metrics that indicate better performance with higher values, we first calculate the average RadCliQ-v1 score for each model across the dataset and then take its reciprocal (1/RadCliQ-v1).} is a composite metric designed to better align with radiologists by combining multiple signals (including clinically grounded graph-based overlap) into a single quality estimate. 
\textbf{RaTEScore} \citep{zhao2024ratescore} is inspired by the RadGraph metric but performs radiology-entity semantic matching with embedding-based similarity (e.g., cosine similarity) under an F1-like formulation, making it robust to paraphrases and synonyms while remaining sensitive to negation.

\subsection{Results}

\paragraph{\textbf{Medical visual question answering:}} Table \ref{tab:vqa} evaluates \textsc{\textsc{MedGPT-oss}}-20B against state-of-the-art open-weight models. Crucially, among these datasets, only SLAKE represents in-domain data; \emph{all other benchmarks strictly evaluate out-of-distribution generalization.}

Despite its smaller 20B parameter footprint and the challenge of unseen distributions, \textsc{\textsc{MedGPT-oss}}-20B establishes new state-of-the-art (SOTA) results on multiple OOD benchmarks. It dominates MedFrameQA (63.01\%), MMMU-dev (61.49\%), and completely eclipses the 32B models on MedXQA (49.23\%, a massive +14.88\% absolute gain over the next best model).

On the benchmarks where it does not claim first place, \textsc{\textsc{MedGPT-oss}}-20B still effectively secures second place with only fractional performance reductions compared to the leading 32B model (Lingshu-32B). For instance, the gap to SOTA is a negligible -0.33\% on MMMU-Med-Pro (4 opt) and -0.71\% on the in-domain SLAKE. This is particularly impressive given Lingshu-32B's explicit training advantage on the MMMU-Med development set.

\begin{table*}[htbp]
\centering
\caption{VQA results on multiple-choice benchmarks. Scores are reported in accuracy (\%).}
\label{tab:vqa}
\resizebox{\textwidth}{!}{
\begin{tabular}{lcccccc}
\toprule
Dataset 
& \textsc{\textsc{MedGPT-oss}}
& OctoMed
& Hulu-Med
& Lingshu$^\dagger$
& MedGemma
& QoQ-Med \\
\midrule \midrule
MedXQA (multimodal)       & \textbf{49.23} & 25.60 & 34.35 & 31.43 & 30.90 & 29.64 \\
SLAKE        & 71.53 & 65.07 & 69.14 & \textbf{72.24} & 55.98 & 46.53 \\
MedFrameQA         & \textbf{63.01} & 42.82 & 62.82 & 61.01 & 47.63 & 55.73 \\
MMMU-Med (dev)          & \textbf{61.49} & 47.65 & 57.71 & 59.43 & 47.43 & 51.84 \\
MMMU-Med-Pro (4 opt)$^*$   & 52.34 & 44.62 & 52.45 & \textbf{52.67} & 45.80 & 46.93 \\
MMMU-Med-Pro (10 opt)    & 39.94 & 23.07 & 37.41 & \textbf{43.45} & 36.71 & 38.12 \\
\bottomrule
\multicolumn{7}{l}{\footnotesize $^\dagger$ Lingshu trained on the MMMU-Med dev set; $^*$opt = options.} \\
\end{tabular}
}
\end{table*}

\paragraph{\textbf{Medical text-only question-answering:}} Table \ref{tab:textqa} details the evaluation of \textsc{\textsc{\textsc{MedGPT-oss}}}-20B across standard text-only medical question-answering benchmarks.  We compare our model against the same suite of open-weight and proprietary models to establish its baseline clinical knowledge and text-based reasoning capabilities.

Despite its smaller 20B parameter footprint, \textsc{\textsc{MedGPT-oss}}-20B achieves new state-of-the-art results on the two most reasoning-intensive benchmarks in our evaluation suite. It claims first place on Medbullets (68.71\%), a dataset specifically designed to emulate challenging, real-world USMLE Step 2 \& 3 clinical scenarios. More notably, it achieves SOTA on the notoriously difficult MedXQA benchmark (25.38\%), successfully outperforming all 32B and 27B models on tasks requiring deep, multi-step medical logic.

On foundational knowledge retrieval benchmarks, \textsc{\textsc{MedGPT-oss}}-20B remains highly competitive, maintaining a tight margin against significantly larger models. On MedQA (70.11\%) and MMLU-Med (72.59\%), it trails the leading 32B models (Hulu-Med and Lingshu) but consistently outperforms the comparably sized MedGemma-27B. This demonstrates that \textsc{\textsc{MedGPT-oss}}-20B successfully preserves broad medical factual recall while simultaneously optimizing for the complex clinical reasoning required to excel on datasets like MedXQA.

\begin{table*}[htbp]
\centering
\caption{Text-only medical QA results. Scores are accuracy (\%).}
\label{tab:textqa}
\resizebox{0.95\textwidth}{!}{
\begin{tabular}{lcccccc}
\toprule
Dataset 
& \textsc{\textsc{MedGPT-oss}}
& OctoMed
& Hulu-Med
& Lingshu
& MedGemma
& QoQ-Med \\
\midrule \midrule
MedQA        & 70.11 & 44.97 & \textbf{77.18} & 70.71 & 66.75 & 55.25 \\
PubMedQA     & 57.81 & 48.31 & 61.00 & \textbf{62.44} & 55.80 & 42.80 \\
MedMCQA      & 62.53    & 55.32 & \textbf{72.75} & 65.27 & 65.48 & 51.42 \\
MedXQA (text)       & \textbf{25.38} & 10.86 & 23.47 & 21.47 & 14.37 & 8.78 \\
MMLU-Med     & 72.59 & 61.65 & \textbf{87.10} & 82.68 & 80.65 & 74.98 \\
Medbullets   & \textbf{68.71} & 32.21 & 67.45 & 58.69 & 51.34 & 37.25 \\
\bottomrule
\end{tabular}
}
\end{table*}

\begin{figure}[htbp]
    \centering
    \includegraphics[width=0.95\linewidth]{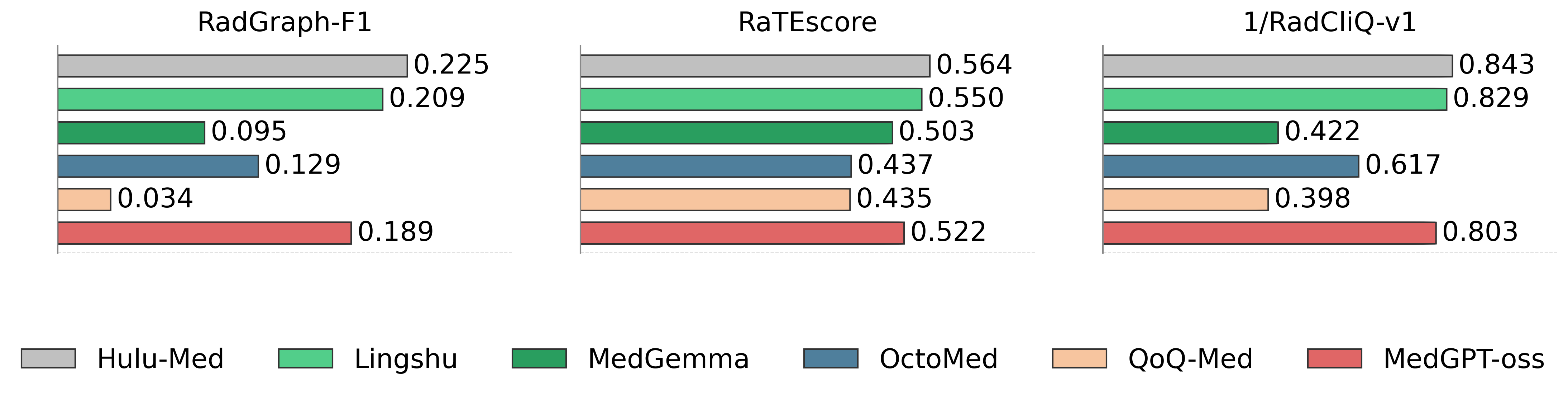}
    \caption{Evaluation of multi-view and longitudinal chest X-ray report generation on the MIMIC-CXR benchmark. Performance is measured across three clinically grounded metrics.}
    \label{fig:report_gen_results}
\end{figure}

\paragraph{\textbf{Chest X-ray report generation:}} Figure \ref{fig:report_gen_results} illustrates the performance of \textsc{MedGPT-oss}-20B on the MIMIC-CXR benchmark compared to baselines. While the larger 32B-parameter models (Hulu-Med and Lingshu) establish the upper bound for this long-form generation task, \textsc{MedGPT-oss}-20B demonstrates highly competitive performance, firmly securing the third position across all three metrics. Specifically, it achieves a RadGraph-F1 of 0.189, a RaTEScore of 0.522, and a 1/RadCliQ-v1 score of 0.803. Notably, \textsc{MedGPT-oss} maintains a narrow performance gap with the 32B models while substantially outperforming other comparably sized or open-weight alternatives. For instance, it roughly doubles the RadGraph-F1 score of MedGemma-27B (0.095) and significantly exceeds OctoMed (0.129). These results highlight that our 20B model successfully preserves high clinical fidelity and minimizes entity-level omissions during complex multimodal generation.

\paragraph{\textbf{In-context Learning Ability.}}
Table~\ref{tab:fewshot} presents a systematic evaluation of in-context learning (ICL) and clinically grounded matching capabilities across the Patient-trial and Impression benchmarks. In the zero-shot setting, \textsc{\textsc{MedGPT-oss}} establishes a strong baseline, achieving state-of-the-art performance on the Impression generation task (47.22\%).

The distinct advantage of \textsc{\textsc{MedGPT-oss}} becomes evident in few-shot evaluation, where ICL inherently measures a model's capacity to leverage provided demonstrations for solving unseen, out-of-distribution tasks. Given a single in-context example on Patient-trial, \textsc{\textsc{MedGPT-oss}} effectively assimilates the contextual signal, yielding a substantial performance gain from 48.81\% (0-shot) to a state-of-the-art 55.60\% (1-shot). Notably, the model processes these raw multimodal examples directly without any auxiliary visual guidance, underscoring the robustness of its native multimodal reasoning.

This strong context utilization stands in stark contrast to competing baselines, which frequently exhibit negative transfer upon the introduction of additional demonstrations. Specifically, Hulu-Med regresses from 51.01\% to 47.00\% on Patient-trial, and Lingshu similarly declines from 43.80\% to 40.27\% on Impression when transitioning from zero-shot to one-shot regimes; OctoMed exhibits analogous degradation across multiple benchmarks. These results suggest that while baseline models are actively distracted by rather than benefiting from extended context, \textsc{\textsc{MedGPT-oss}} demonstrates a genuinely superior capacity to integrate and reason over novel contextual evidence for OOD clinical tasks.

\begin{table*}[h]
\centering
\caption{Few-shot benchmarks. Scores are reported as accuracy percentages (\%), with the exception of the Impression benchmark, which is evaluated using RaTEScore.}
\label{tab:fewshot}
\resizebox{\textwidth}{!}{
\begin{tabular}{lcccccc}
\toprule
Dataset 
& \textsc{\textsc{MedGPT-oss}}
& OctoMed
& Hulu-Med
& Lingshu
& MedGemma
& QoQ-Med \\
\midrule \midrule
Patient-trial (0-shot) & 48.81 & 40.96 & 51.01 & \textbf{52.07} & 31.03 & 45.20 \\
Patient-trial (1-shot) & \textbf{55.60} & 40.02 & 47.00 & 48.91 & 52.24 & 47.41 \\
Impression (0-shot) & \textbf{47.22} & 31.04 & 43.14 & 43.80 & 38.42 & 41.44 \\
Impression (1-shot) & \textbf{47.25}   & 30.91 & 41.52 & 40.27 & 38.29 & 40.71 \\
\bottomrule
\end{tabular}
}
\end{table*}

\section{Discussion}

Our current results indicate that a modular medical VLM recipe, a strong open language model (GPT-oss) coupled with a lightweight vision front-end, can achieve competitive performance across heterogeneous biomedical tasks when paired with a staged training curriculum.

Despite these promising results, several critical limitations remain. Primarily, \textsc{MedGPT-oss} is positioned as a research foundation rather than a drop-in clinical tool. Safe real-world deployment necessitates rigorous external validation to account for domain shifts across different healthcare institutions and imaging hardware. Additionally, while our model shows strong factual grounding, the generation of long-form clinical text (such as radiology reports) remains susceptible to subtle hallucinations or omissions of complex findings, a persistent challenge in medical VLMs. Finally, continuous monitoring and auditing are required to address fairness and mitigate inherent biases present in the underlying training corpora.

To address these limitations and advance the capabilities of generalist biomedical AI, we outline several key directions for future development:

\textbf{Scaling Model Parameters:} While our 20B configuration offers a strong balance of performance and deployability, we plan to systematically scale the underlying language backbone and visual encoders. Exploring larger dense architectures could unlock deeper clinical reasoning and broader zero-shot and few-shot generalization \citep{kaplan2020scaling}.

\textbf{Incorporating Native 3D Image Processing:} Currently, \textsc{MedGPT-oss} relies on 2D visual feature extraction. Clinical decision-making frequently depends on volumetric imaging, such as CT and MRI scans \citep{wasserthal2023totalsegmentator, zhao2024normative}. We aim to integrate native 3D vision encoders to process spatial depth and temporal sequences, providing a more holistic understanding of complex radiological studies.

\textbf{Reinforcement Learning with Clinical Rubrics:} To further align report generation with expert radiologist standards, we plan to implement reinforcement learning (e.g., RLHF or RLAIF) \citep{ouyang2022training, bai2022constitutional}. By utilizing clinically informed rubrics and metrics as reward signals, we can directly penalize critical omissions, incorrect negations, and clinical hallucinations \citep{zhang2025editgrpo}.

\textbf{Agentic Ability Training:} Clinical workflows are inherently interactive. We aim to transition \textsc{MedGPT-oss} from a passive question-answering system into a domain-specific clinical agent \citep{zhang2026radagents}. By training the model to utilize external tools (e.g., querying electronic health records, executing medical calculators, or retrieving literature), it could support complex, multi-step clinical reasoning and longitudinal patient management \citep{team2026kimi}.

\section*{Acknowledgements}

We gratefully acknowledge the University of Florida Research Computing for providing the HiPerGator supercomputing infrastructure and technical support that enabled the research reported in this work. This study was partially supported by grants from the PARADIGM program awarded by the Advanced Research Projects Agency for Health (ARPA-H). Lichao Sun was partially supported by the National Science Foundation under Grants CRII-2246067, ATD-2427915, NSF POSE-2346158, and NSF POSE-2449280, as well as by Lehigh University under Grant FRGS00011497.

\bibliography{iclr2026_conference}
\bibliographystyle{iclr2026_conference}


\end{document}

%% file: math_commands.tex
\usepackage{amsmath,amsfonts,bm}

\def\eqref#1{equation~\ref{#1}}

\def\1{\bm{1}}

\DeclareMathAlphabet{\mathsfit}{\encodingdefault}{\sfdefault}{m}{sl}
\SetMathAlphabet{\mathsfit}{bold}{\encodingdefault}{\sfdefault}{bx}{n}



%% file: iclr2026_conference.bib
@article{thirunavukarasu2023large,
  title={Large language models in medicine},
  author={Thirunavukarasu, Arun James and Ting, Darren Shu Jeng and Elangovan, Kabilan and Gutierrez, Laura and Tan, Ting Fang and Ting, Daniel Shu Wei},
  journal={Nature medicine},
  volume={29},
  number={8},
  pages={1930--1940},
  year={2023},
  publisher={Nature Publishing Group US New York}
}

@article{moor2023foundation,
  title={Foundation models for generalist medical artificial intelligence},
  author={Moor, Michael and Banerjee, Oishi and Abad, Zahra Shakeri Hossein and Krumholz, Harlan M and Leskovec, Jure and Topol, Eric J and Rajpurkar, Pranav},
  journal={Nature},
  volume={616},
  number={7956},
  pages={259--265},
  year={2023},
  publisher={Nature Publishing Group UK London}
}

@article{zhang2024generalist,
  title={A generalist vision--language foundation model for diverse biomedical tasks},
  author={Zhang, Kai and Zhou, Rong and Adhikarla, Eashan and Yan, Zhiling and Liu, Yixin and Yu, Jun and Liu, Zhengliang and Chen, Xun and Davison, Brian D and Ren, Hui and others},
  journal={Nature Medicine},
  volume={30},
  number={11},
  pages={3129--3141},
  year={2024},
  publisher={Nature Publishing Group US New York}
}

@article{zhang2025multimodal,
  title={A multimodal biomedical foundation model trained from fifteen million image--text pairs},
  author={Zhang, Sheng and Xu, Yanbo and Usuyama, Naoto and Xu, Hanwen and Bagga, Jaspreet and Tinn, Robert and Preston, Sam and Rao, Rajesh and Wei, Mu and Valluri, Naveen and others},
  journal={Nejm Ai},
  volume={2},
  number={1},
  pages={AIoa2400640},
  year={2025},
  publisher={Massachusetts Medical Society}
}

@misc{llava_med,
  title = {LLaVA-Med: Training a Large Language-and-Vision Assistant for Biomedicine in One Day},
  author = {Haotian Liu and Qing Liu and Amit D. Sudarshan and Mert R. Sabuncu and Marinka Zitnik and Li Fei-Fei and Jure Leskovec and Bo Wang},
  year = {2023},
  eprint = {2310.03744},
  archivePrefix = {arXiv},
  primaryClass = {cs.CL},
  url = {https://arxiv.org/abs/2310.03744}
}

@article{agarwal2025gpt,
  title={gpt-oss-120b \& gpt-oss-20b Model Card},
  author={Agarwal, Sandhini and Ahmad, Lama and Ai, Jason and Altman, Sam and Applebaum, Andy and Arbus, Edwin and Arora, Rahul K and Bai, Yu and Baker, Bowen and Bao, Haiming and others},
  journal={arXiv preprint arXiv:2508.10925},
  year={2025}
}

@article{Lau2018VQARAD,
  title   = {A Dataset of Clinically Generated Visual Questions and Answers About Radiology Images},
  author  = {Lau, Jason J. and Gayen, Soumya and Ben Abacha, Asma and Demner-Fushman, Dina},
  journal = {Scientific Data},
  volume  = {5},
  pages   = {180251},
  year    = {2018},
  doi     = {10.1038/sdata.2018.251}
}

@inproceedings{Liu2021SLAKE,
  title     = {SLAKE: A Semantically Labeled Knowledge-Enhanced Dataset for Medical Visual Question Answering},
  author    = {Liu, Bo and Zhan, Li-Ming and Xu, Li and Ma, Lin and Yang, Yan and Wu, Xiao-Ming},
  booktitle = {2021 IEEE 18th International Symposium on Biomedical Imaging (ISBI)},
  year      = {2021},
  pages     = {1--5},
  doi       = {10.1109/ISBI48211.2021.9433885}
}

@article{Lu2021CLAM,
  title   = {Data-efficient and Weakly Supervised Computational Pathology on Whole-slide Images},
  author  = {Lu, Ming Y. and Williamson, Drew F. K. and Chen, Tiffany Y. and Chen, Richard J. and Barbieri, Matteo and Mahmood, Faisal},
  journal = {Nature Biomedical Engineering},
  year    = {2021},
  doi     = {10.1038/s41551-020-00682-w}
}

@inproceedings{Romanov2018MedNLI,
  title     = {Lessons from Natural Language Inference in the Clinical Domain},
  author    = {Romanov, Alexey and Shivade, Chaitanya},
  booktitle = {Proceedings of the 2018 Conference on Empirical Methods in Natural Language Processing (EMNLP)},
  year      = {2018},
  pages     = {1586--1596},
  doi       = {10.18653/v1/D18-1187}
}

@article{Tu2023,
  author       = {Tao Tu and Shekoofeh Azizi and Danny Driess and Mike Schaekermann and Mohamed Amin and Pi{-}Chuan Chang and Andrew Carroll and Chuck Lau and Ryutaro Tanno and Ira Ktena and Basil Mustafa and Aakanksha Chowdhery and Yun Liu and Simon Kornblith and David Fleet and Philip Mansfield and Sushant Prakash and Renee Wong and Sunny Virmani and Christopher Semturs and S.~Sara Mahdavi and Bradley Green and Ewa Dominowska and Blaise {Aguera y Arcas} and Joelle Barral and Dale Webster and Gregory~S. Corrado and Yossi Matias and Karan Singhal and Pete Florence and Alan Karthikesalingam and Vivek Natarajan},
  title        = {Towards Generalist Biomedical {AI}},
  journal      = {arXiv preprint arXiv:2307.14334},
  year         = {2023}
}

@article{OpenAI2023,
  author       = {{OpenAI}},
  title        = {{GPT-4} Technical Report},
  journal      = {arXiv preprint arXiv:2303.08774},
  year         = {2023}
}

@article{yao2024deco,
  title={Deco: Decoupling token compression from semantic abstraction in multimodal large language models},
  author={Yao, Linli and Li, Lei and Ren, Shuhuai and Wang, Lean and Liu, Yuanxin and Sun, Xu and Hou, Lu},
  journal={arXiv preprint arXiv:2405.20985},
  year={2024}
}

@inproceedings{li2023blip,
  title={Blip-2: Bootstrapping language-image pre-training with frozen image encoders and large language models},
  author={Li, Junnan and Li, Dongxu and Savarese, Silvio and Hoi, Steven},
  booktitle={International conference on machine learning},
  pages={19730--19742},
  year={2023},
  organization={PMLR}
}

@inproceedings{carion2020end,
  title={End-to-end object detection with transformers},
  author={Carion, Nicolas and Massa, Francisco and Synnaeve, Gabriel and Usunier, Nicolas and Kirillov, Alexander and Zagoruyko, Sergey},
  booktitle={European conference on computer vision},
  pages={213--229},
  year={2020},
  organization={Springer}
}

@article{liu2023visual,
  title={Visual instruction tuning},
  author={Liu, Haotian and Li, Chunyuan and Wu, Qingyang and Lee, Yong Jae},
  journal={Advances in neural information processing systems},
  volume={36},
  pages={34892--34916},
  year={2023}
}

@article{yuan2023tinygpt,
  title={Tinygpt-v: Efficient multimodal large language model via small backbones},
  author={Yuan, Zhengqing and Li, Zhaoxu and Huang, Weiran and Ye, Yanfang and Sun, Lichao},
  journal={arXiv preprint arXiv:2312.16862},
  year={2023}
}

@article{arora2025healthbench,
  title={Healthbench: Evaluating large language models towards improved human health},
  author={Arora, Rahul K and Wei, Jason and Hicks, Rebecca Soskin and Bowman, Preston and Qui{\~n}onero-Candela, Joaquin and Tsimpourlas, Foivos and Sharman, Michael and Shah, Meghan and Vallone, Andrea and Beutel, Alex and others},
  journal={arXiv preprint arXiv:2505.08775},
  year={2025}
}

@inproceedings{radford2021learning,
  title={Learning transferable visual models from natural language supervision},
  author={Radford, Alec and Kim, Jong Wook and Hallacy, Chris and Ramesh, Aditya and Goh, Gabriel and Agarwal, Sandhini and Sastry, Girish and Askell, Amanda and Mishkin, Pamela and Clark, Jack and others},
  booktitle={International conference on machine learning},
  pages={8748--8763},
  year={2021},
  organization={PmLR}
}

@article{xu2025lingshu,
  title={Lingshu: A Generalist Foundation Model for Unified Multimodal Medical Understanding and Reasoning},
  author={Xu, Weiwen and Chan, Hou Pong and Li, Long and Aljunied, Mahani and Yuan, Ruifeng and Wang, Jianyu and Xiao, Chenghao and Chen, Guizhen and Liu, Chaoqun and Li, Zhaodonghui and others},
  journal={arXiv preprint arXiv:2506.07044},
  year={2025}
}

@article{jiang2025hulu,
  title={Hulu-med: A transparent generalist model towards holistic medical vision-language understanding},
  author={Jiang, Songtao and Wang, Yuan and Song, Sibo and Hu, Tianxiang and Zhou, Chenyi and Pu, Bin and Zhang, Yan and Yang, Zhibo and Feng, Yang and Zhou, Joey Tianyi and others},
  journal={arXiv preprint arXiv:2510.08668},
  year={2025}
}

@inproceedings{chen-etal-2024-towards-injecting,
    title = "Towards Injecting Medical Visual Knowledge into Multimodal {LLM}s at Scale",
    author = "Chen, Junying  and
      Gui, Chi  and
      Ouyang, Ruyi  and
      Gao, Anningzhe  and
      Chen, Shunian  and
      Chen, Guiming Hardy  and
      Wang, Xidong  and
      Cai, Zhenyang  and
      Ji, Ke  and
      Wan, Xiang  and
      Wang, Benyou",
    editor = "Al-Onaizan, Yaser  and
      Bansal, Mohit  and
      Chen, Yun-Nung",
    booktitle = "Proceedings of the 2024 Conference on Empirical Methods in Natural Language Processing",
    month = nov,
    year = "2024",
    address = "Miami, Florida, USA",
    publisher = "Association for Computational Linguistics",
    url = "https://aclanthology.org/2024.emnlp-main.418/",
    doi = "10.18653/v1/2024.emnlp-main.418",
    pages = "7346--7370"
}

@article{an2025llava,
  title={Llava-onevision-1.5: Fully open framework for democratized multimodal training},
  author={An, Xiang and Xie, Yin and Yang, Kaicheng and Zhang, Wenkang and Zhao, Xiuwei and Cheng, Zheng and Wang, Yirui and Xu, Songcen and Chen, Changrui and Zhu, Didi and others},
  journal={arXiv preprint arXiv:2509.23661},
  year={2025}
}

@article{bai2023qwen,
  title={Qwen technical report},
  author={Bai, Jinze and Bai, Shuai and Chu, Yunfei and Cui, Zeyu and Dang, Kai and Deng, Xiaodong and Fan, Yang and Ge, Wenbin and Han, Yu and Huang, Fei and others},
  journal={arXiv preprint arXiv:2309.16609},
  year={2023}
}

@article{Qwen3-VL,
      title={Qwen3-VL Technical Report}, 
      author={Shuai Bai and Yuxuan Cai and Ruizhe Chen and Keqin Chen and Xionghui Chen and Zesen Cheng and Lianghao Deng and Wei Ding and Chang Gao and Chunjiang Ge and Wenbin Ge and Zhifang Guo and Qidong Huang and Jie Huang and Fei Huang and Binyuan Hui and Shutong Jiang and Zhaohai Li and Mingsheng Li and Mei Li and Kaixin Li and Zicheng Lin and Junyang Lin and Xuejing Liu and Jiawei Liu and Chenglong Liu and Yang Liu and Dayiheng Liu and Shixuan Liu and Dunjie Lu and Ruilin Luo and Chenxu Lv and Rui Men and Lingchen Meng and Xuancheng Ren and Xingzhang Ren and Sibo Song and Yuchong Sun and Jun Tang and Jianhong Tu and Jianqiang Wan and Peng Wang and Pengfei Wang and Qiuyue Wang and Yuxuan Wang and Tianbao Xie and Yiheng Xu and Haiyang Xu and Jin Xu and Zhibo Yang and Mingkun Yang and Jianxin Yang and An Yang and Bowen Yu and Fei Zhang and Hang Zhang and Xi Zhang and Bo Zheng and Humen Zhong and Jingren Zhou and Fan Zhou and Jing Zhou and Yuanzhi Zhu and Ke Zhu},
	  journal={arXiv preprint arXiv:2511.21631},
      year={2025}
}

@inproceedings{
hu2022lora,
title={Lo{RA}: Low-Rank Adaptation of Large Language Models},
author={Edward J Hu and yelong shen and Phillip Wallis and Zeyuan Allen-Zhu and Yuanzhi Li and Shean Wang and Lu Wang and Weizhu Chen},
booktitle={International Conference on Learning Representations},
year={2022},
url={https://openreview.net/forum?id=nZeVKeeFYf9}
}

@inproceedings{zhai2023sigmoid,
  title={Sigmoid loss for language image pre-training},
  author={Zhai, Xiaohua and Mustafa, Basil and Kolesnikov, Alexander and Beyer, Lucas},
  booktitle={Proceedings of the IEEE/CVF international conference on computer vision},
  pages={11975--11986},
  year={2023}
}

@article{sellergren2025medgemma,
  title={Medgemma technical report},
  author={Sellergren, Andrew and Kazemzadeh, Sahar and Jaroensri, Tiam and Kiraly, Atilla and Traverse, Madeleine and Kohlberger, Timo and Xu, Shawn and Jamil, Fayaz and Hughes, C{\'\i}an and Lau, Charles and others},
  journal={arXiv preprint arXiv:2507.05201},
  year={2025}
}

@inproceedings{lin2023pmc,
  title={Pmc-clip: Contrastive language-image pre-training using biomedical documents},
  author={Lin, Weixiong and Zhao, Ziheng and Zhang, Xiaoman and Wu, Chaoyi and Zhang, Ya and Wang, Yanfeng and Xie, Weidi},
  booktitle={International Conference on Medical Image Computing and Computer-Assisted Intervention},
  pages={525--536},
  year={2023},
  organization={Springer}
}

@article{ikezogwo2023quilt,
  title={Quilt-1m: One million image-text pairs for histopathology},
  author={Ikezogwo, Wisdom and Seyfioglu, Saygin and Ghezloo, Fatemeh and Geva, Dylan and Sheikh Mohammed, Fatwir and Anand, Pavan Kumar and Krishna, Ranjay and Shapiro, Linda},
  journal={Advances in neural information processing systems},
  volume={36},
  pages={37995--38017},
  year={2023}
}

@article{ruckert2024rocov2,
  title={Rocov2: Radiology objects in context version 2, an updated multimodal image dataset},
  author={R{\"u}ckert, Johannes and Bloch, Louise and Br{\"u}ngel, Raphael and Idrissi-Yaghir, Ahmad and Sch{\"a}fer, Henning and Schmidt, Cynthia S and Koitka, Sven and Pelka, Obioma and Abacha, Asma Ben and G. Seco de Herrera, Alba and others},
  journal={Scientific Data},
  volume={11},
  number={1},
  pages={688},
  year={2024},
  publisher={Nature Publishing Group UK London}
}

@inproceedings{lozano2025biomedica,
  title={Biomedica: An open biomedical image-caption archive, dataset, and vision-language models derived from scientific literature},
  author={Lozano, Alejandro and Sun, Min Woo and Burgess, James and Chen, Liangyu and Nirschl, Jeffrey J and Gu, Jeffrey and Lopez, Ivan and Aklilu, Josiah and Rau, Anita and Katzer, Austin Wolfgang and others},
  booktitle={Proceedings of the Computer Vision and Pattern Recognition Conference},
  pages={19724--19735},
  year={2025}
}

@article{johnson2019mimic,
  title={MIMIC-CXR, a de-identified publicly available database of chest radiographs with free-text reports},
  author={Johnson, Alistair EW and Pollard, Tom J and Berkowitz, Seth J and Greenbaum, Nathaniel R and Lungren, Matthew P and Deng, Chih-ying and Mark, Roger G and Horng, Steven},
  journal={Scientific data},
  volume={6},
  number={1},
  pages={317},
  year={2019},
  publisher={Nature Publishing Group UK London}
}

@inproceedings{
xie2025medtrinitym,
title={MedTrinity-25M: A Large-scale Multimodal Dataset with Multigranular Annotations for Medicine},
author={Yunfei Xie and Ce Zhou and Lang Gao and Juncheng Wu and Xianhang Li and Hong-Yu Zhou and Sheng Liu and Lei Xing and James Zou and Cihang Xie and Yuyin Zhou},
booktitle={The Thirteenth International Conference on Learning Representations},
year={2025},
url={https://openreview.net/forum?id=IwgmgidYPS}
}

@inproceedings{zhang2025editgrpo,
  title={Editgrpo: Reinforcement learning with post-rollout edits for clinically accurate chest x-ray report generation},
  author={Zhang, Kai and Malon, Christopher and Sun, Lichao and Min, Martin Renqiang},
  booktitle={Proceedings of the 14th International Joint Conference on Natural Language Processing and the 4th Conference of the Asia-Pacific Chapter of the Association for Computational Linguistics},
  pages={304--316},
  year={2025}
}

@inproceedings{zhang2025libra,
  title={Libra: Leveraging temporal images for biomedical radiology analysis},
  author={Zhang, Xi and Meng, Zaiqiao and Lever, Jake and Ho, Edmond SL},
  booktitle={Findings of the Association for Computational Linguistics: ACL 2025},
  pages={17275--17303},
  year={2025}
}

@article{yu2025medframeqa,
  title={Medframeqa: A multi-image medical vqa benchmark for clinical reasoning},
  author={Yu, Suhao and Wang, Haojin and Wu, Juncheng and Luo, Luyang and Wang, Jingshen and Xie, Cihang and Rajpurkar, Pranav and Yang, Carl and Yang, Yang and Wang, Kang and others},
  journal={arXiv preprint arXiv:2505.16964},
  year={2025}
}

@inproceedings{yue2024mmmu,
  title={Mmmu: A massive multi-discipline multimodal understanding and reasoning benchmark for expert agi},
  author={Yue, Xiang and Ni, Yuansheng and Zhang, Kai and Zheng, Tianyu and Liu, Ruoqi and Zhang, Ge and Stevens, Samuel and Jiang, Dongfu and Ren, Weiming and Sun, Yuxuan and others},
  booktitle={Proceedings of the IEEE/CVF conference on computer vision and pattern recognition},
  pages={9556--9567},
  year={2024}
}

@inproceedings{yue2025mmmu,
  title={Mmmu-pro: A more robust multi-discipline multimodal understanding benchmark},
  author={Yue, Xiang and Zheng, Tianyu and Ni, Yuansheng and Wang, Yubo and Zhang, Kai and Tong, Shengbang and Sun, Yuxuan and Yu, Botao and Zhang, Ge and Sun, Huan and others},
  booktitle={Proceedings of the 63rd Annual Meeting of the Association for Computational Linguistics (Volume 1: Long Papers)},
  pages={15134--15186},
  year={2025}
}

@inproceedings{zuo2025medxpertqa,
  title={MedXpertQA: Benchmarking Expert-Level Medical Reasoning and Understanding},
  author={Zuo, Yuxin and Qu, Shang and Li, Yifei and Chen, Zhang-Ren and Zhu, Xuekai and Hua, Ermo and Zhang, Kaiyan and Ding, Ning and Zhou, Bowen},
  booktitle={International Conference on Machine Learning},
  pages={80961--80990},
  year={2025},
  organization={PMLR}
}

@article{jin2021disease,
  title={What disease does this patient have? a large-scale open domain question answering dataset from medical exams},
  author={Jin, Di and Pan, Eileen and Oufattole, Nassim and Weng, Wei-Hung and Fang, Hanyi and Szolovits, Peter},
  journal={Applied Sciences},
  volume={11},
  number={14},
  pages={6421},
  year={2021},
  publisher={MDPI}
}

@inproceedings{zhao2024ratescore,
  title={Ratescore: A metric for radiology report generation},
  author={Zhao, Weike and Wu, Chaoyi and Zhang, Xiaoman and Zhang, Ya and Wang, Yanfeng and Xie, Weidi},
  booktitle={Proceedings of the 2024 Conference on Empirical Methods in Natural Language Processing},
  pages={15004--15019},
  year={2024}
}

@inproceedings{
ye2025justice,
title={Justice or Prejudice? Quantifying Biases in {LLM}-as-a-Judge},
author={Jiayi Ye and Yanbo Wang and Yue Huang and Dongping Chen and Qihui Zhang and Nuno Moniz and Tian Gao and Werner Geyer and Chao Huang and Pin-Yu Chen and Nitesh V Chawla and Xiangliang Zhang},
booktitle={The Thirteenth International Conference on Learning Representations},
year={2025},
url={https://openreview.net/forum?id=3GTtZFiajM}
}

@inproceedings{jin2019pubmedqa,
  title={Pubmedqa: A dataset for biomedical research question answering},
  author={Jin, Qiao and Dhingra, Bhuwan and Liu, Zhengping and Cohen, William and Lu, Xinghua},
  booktitle={Proceedings of the 2019 conference on empirical methods in natural language processing and the 9th international joint conference on natural language processing (EMNLP-IJCNLP)},
  pages={2567--2577},
  year={2019}
}

@inproceedings{pal2022medmcqa,
  title={Medmcqa: A large-scale multi-subject multi-choice dataset for medical domain question answering},
  author={Pal, Ankit and Umapathi, Logesh Kumar and Sankarasubbu, Malaikannan},
  booktitle={Conference on health, inference, and learning},
  pages={248--260},
  year={2022},
  organization={PMLR}
}

@inproceedings{
hendrycks2021measuring,
title={Measuring Massive Multitask Language Understanding},
author={Dan Hendrycks and Collin Burns and Steven Basart and Andy Zou and Mantas Mazeika and Dawn Song and Jacob Steinhardt},
booktitle={International Conference on Learning Representations},
year={2021},
url={https://openreview.net/forum?id=d7KBjmI3GmQ}
}

@inproceedings{chen2025benchmarking,
  title={Benchmarking large language models on answering and explaining challenging medical questions},
  author={Chen, Hanjie and Fang, Zhouxiang and Singla, Yash and Dredze, Mark},
  booktitle={Proceedings of the 2025 Conference of the Nations of the Americas Chapter of the Association for Computational Linguistics: Human Language Technologies (Volume 1: Long Papers)},
  pages={3563--3599},
  year={2025}
}

@inproceedings{roberts2022overview,
  title={Overview of the TREC 2022 Clinical Trials Track.},
  author={Roberts, Kirk and Demner-Fushman, Dina and Voorhees, Ellen M and Bedrick, Steven and Hersh, William R},
  booktitle={TREC},
  year={2022}
}

@inproceedings{soboroff2021overview,
  title={Overview of TREC 2021.},
  author={Soboroff, Ian},
  booktitle={TREC},
  year={2021}
}

@inproceedings{koopman2016test,
  title={A test collection for matching patients to clinical trials},
  author={Koopman, Bevan and Zuccon, Guido},
  booktitle={Proceedings of the 39th International ACM SIGIR conference on Research and Development in Information Retrieval},
  pages={669--672},
  year={2016}
}

@article{ossowski2025octomed,
  title={OctoMed: Data Recipes for State-of-the-Art Multimodal Medical Reasoning},
  author={Ossowski, Timothy and Zhang, Sheng and Liu, Qianchu and Qin, Guanghui and Tan, Reuben and Naumann, Tristan and Hu, Junjie and Poon, Hoifung},
  journal={arXiv preprint arXiv:2511.23269},
  year={2025}
}

@inproceedings{
dai2025qoqmed,
title={QoQ-Med: Building Multimodal Clinical Foundation Models with Domain-Aware {GRPO} Training},
author={Wei Dai and Peilin Chen and Chanakya Ekbote and Paul Pu Liang},
booktitle={The Thirty-ninth Annual Conference on Neural Information Processing Systems},
year={2025},
url={https://openreview.net/forum?id=ZwCVFBFUFb}
}

@article{yu2023evaluating,
  title={Evaluating progress in automatic chest x-ray radiology report generation},
  author={Yu, Feiyang and Endo, Mark and Krishnan, Rayan and Pan, Ian and Tsai, Andy and Reis, Eduardo Pontes and Fonseca, Eduardo Kaiser Ururahy Nunes and Lee, Henrique Min Ho and Abad, Zahra Shakeri Hossein and Ng, Andrew Y and others},
  journal={Patterns},
  volume={4},
  number={9},
  year={2023},
  publisher={Elsevier}
}

@inproceedings{
jain2021radgraph,
title={RadGraph: Extracting Clinical Entities and Relations from Radiology Reports},
author={Saahil Jain and Ashwin Agrawal and Adriel Saporta and Steven Truong and Du Nguyen Duong and Tan Bui and Pierre Chambon and Yuhao Zhang and Matthew P. Lungren and Andrew Y. Ng and Curtis Langlotz and Pranav Rajpurkar},
booktitle={Thirty-fifth Conference on Neural Information Processing Systems Datasets and Benchmarks Track (Round 1)},
year={2021},
url={https://openreview.net/forum?id=pMWtc5NKd7V}
}

@inproceedings{
zhang2026radagents,
title={RadAgents: Multimodal Agentic Reasoning for Chest X-ray Interpretation with Radiologist-like Workflows},
author={Kai Zhang and Corey D Barrett and Jangwon Kim and Lichao Sun and Tara Taghavi and Krishnaram Kenthapadi},
booktitle={Medical Imaging with Deep Learning},
year={2026},
url={https://openreview.net/forum?id=2mlxx8R0Ru}
}

@article{wasserthal2023totalsegmentator,
  title={TotalSegmentator: robust segmentation of 104 anatomic structures in CT images},
  author={Wasserthal, Jakob and Breit, Hanns-Christian and Meyer, Manfred T and Pradella, Maurice and Hinck, Daniel and Sauter, Alexander W and Heye, Tobias and Boll, Daniel T and Cyriac, Joshy and Yang, Shan and others},
  journal={Radiology: Artificial Intelligence},
  volume={5},
  number={5},
  pages={e230024},
  year={2023},
  publisher={Radiological Society of North America}
}

@inproceedings{zhao2024normative,
  title={Normative modeling with focal loss and adversarial autoencoders for alzheimer’s disease diagnosis and biomarker identification},
  author={Zhao, Songlin and Zhou, Rong and Zhang, Yu and Chen, Yong and He, Lifang},
  booktitle={International workshop on applications of medical AI},
  pages={231--240},
  year={2024},
  organization={Springer}
}

@article{yang2022large,
  title={A large language model for electronic health records},
  author={Yang, Xi and Chen, Aokun and PourNejatian, Nima and Shin, Hoo Chang and Smith, Kaleb E and Parisien, Christopher and Compas, Colin and Martin, Cheryl and Costa, Anthony B and Flores, Mona G and others},
  journal={NPJ digital medicine},
  volume={5},
  number={1},
  pages={194},
  year={2022},
  publisher={Nature Publishing Group UK London}
}

@article{peng2025scaling,
  title={Scaling up biomedical vision-language models: Fine-tuning, instruction tuning, and multi-modal learning},
  author={Peng, Cheng and Zhang, Kai and Lyu, Mengxian and Liu, Hongfang and Sun, Lichao and Wu, Yonghui},
  journal={Journal of Biomedical Informatics},
  pages={104946},
  year={2025},
  publisher={Elsevier}
}

@article{team2026kimi,
  title={Kimi K2. 5: Visual Agentic Intelligence},
  author={Team, Kimi and Bai, Tongtong and Bai, Yifan and Bao, Yiping and Cai, SH and Cao, Yuan and Charles, Y and Che, HS and Chen, Cheng and Chen, Guanduo and others},
  journal={arXiv preprint arXiv:2602.02276},
  year={2026}
}

@article{ouyang2022training,
  title={Training language models to follow instructions with human feedback},
  author={Ouyang, Long and Wu, Jeffrey and Jiang, Xu and Almeida, Diogo and Wainwright, Carroll and Mishkin, Pamela and Zhang, Chong and Agarwal, Sandhini and Slama, Katarina and Ray, Alex and others},
  journal={Advances in neural information processing systems},
  volume={35},
  pages={27730--27744},
  year={2022}
}

@article{bai2022constitutional,
  title={Constitutional ai: Harmlessness from ai feedback},
  author={Bai, Yuntao and Kadavath, Saurav and Kundu, Sandipan and Askell, Amanda and Kernion, Jackson and Jones, Andy and Chen, Anna and Goldie, Anna and Mirhoseini, Azalia and McKinnon, Cameron and others},
  journal={arXiv preprint arXiv:2212.08073},
  year={2022}
}

@article{kaplan2020scaling,
  title={Scaling laws for neural language models},
  author={Kaplan, Jared and McCandlish, Sam and Henighan, Tom and Brown, Tom B and Chess, Benjamin and Child, Rewon and Gray, Scott and Radford, Alec and Wu, Jeffrey and Amodei, Dario},
  journal={arXiv preprint arXiv:2001.08361},
  year={2020}
}

@inproceedings{rasley2020deepspeed,
  title={Deepspeed: System optimizations enable training deep learning models with over 100 billion parameters},
  author={Rasley, Jeff and Rajbhandari, Samyam and Ruwase, Olatunji and He, Yuxiong},
  booktitle={Proceedings of the 26th ACM SIGKDD international conference on knowledge discovery \& data mining},
  pages={3505--3506},
  year={2020}
}

@inproceedings{
loshchilov2018decoupled,
title={Decoupled Weight Decay Regularization},
author={Ilya Loshchilov and Frank Hutter},
booktitle={International Conference on Learning Representations},
year={2019},
url={https://openreview.net/forum?id=Bkg6RiCqY7},
}

@inproceedings{
peng2024yarn,
title={Ya{RN}: Efficient Context Window Extension of Large Language Models},
author={Bowen Peng and Jeffrey Quesnelle and Honglu Fan and Enrico Shippole},
booktitle={The Twelfth International Conference on Learning Representations},
year={2024},
url={https://openreview.net/forum?id=wHBfxhZu1u}
}
